# Manipulating human-oriented dictionaries with very simple tools

Jean Gaschler & Mathieu Lafourcade
(Jean.Gaschler@imag.fr  -  Mathieu.Lafourcade@imag.fr)

GETA, IMAG-campus (UJF & CNRS)
BP 53, F-38041 GRENOBLE Cedex 09

## Abstract

It is possible to manipulate real-size human-oriented dictionaries on a Macintosh by using only very simple tools. Our methodology has been applied in the construction of a French-English-Malay dictionary. This dictionary has been obtained by "crossing" semi-automatically two bilingual dictionaries. To revise the dictionary, as well as to obtain a publishable paper form and an on-line electronic form, we use only Microsoft Word$^{TM}$, a specialized language for writing transcriptors and a small but powerful dictionary tool.

## Introduction

In collaboration with University Sains Malaysia (USM), we are working on a French-English-Malay human-oriented dictionary (FEM project) obtained by "crossing" French-English and English-Malay dictionaries.

Taking into account the reluctance of lexicographers to revise dictionaries through database interfaces (dBASEIII$^{TM}$ or 4D$^{TM}$), we have developed a methodology based on using only very simple tools. For editing, we use Word and its styling facility, because no editor of structured document is available on the Macintosh. For importing and exporting, we use LT, a simple specialized language for writing transcriptors, and transform between representations (normalized ASCII, RTF, etc.). Finally, we have developed ALEX, a dictionary tool, to support the electronic form. The methods defined have been applied on the FEM dictionary. They concern the correction of errors which can appears in a manually built dictionary and the formatting of this dictionary.

We introduce first in more details the situation we face with the FEM project. We expose the goals we tend to reach. Then we gives our generic methods and their applications to the specific case of the FEM dictionary.

## I. Situation

### 1. Presentation

The FEM dictionary is composed of two parts: a general one (12,000 entries) and a specific one for computer-science terminology (2,300 terms). Both paper and electronic forms will be produced by mid-94.

We have initially received ASCII files obtained firstly by optical characters recognition and corrected manually in which the informations of the French-English and English-Malay dictionaries have been crossed.

### 2. Logical form

The printed form of dictionaries reflect their internal structure (Boguraev 1990, Byrd, Calzolari, Chodorow & al. 1987). This structure can be modelized with a logical form which gives the sequence of the informations contained by the dictionary. This logical form contains entries, prononciation parts, spelling variants, grammatical categories, semantics information, sub-entries, etc. We have defined a logical form of the article in the FEM dictionary (fig. 1).

| | |
|---|---|
| **French entry** (Masculine by default)<br>    Pronunciation<br>    *Entry variant*<br>    *Pronunciation variant*<br>    *Plural form*<br>    *Plural form pronunciation*<br>    *Masculine plural form*<br>    *Masculine plural form pronunciation*<br>    *Masc plural form pronunciation variant*<br>    *Feminine form*<br>    *Feminine form pronunciation*<br>    *Feminine plural form*<br>    *Feminine plural form pronunciation*<br>    *Plural form variant*<br>    *Plural form variant pronunciation*<br>    French grammar category<br>        *Gloss (in French)*<br>            (semantic categories)<br>        *Label English equivalent*<br>        English equivalent<br>        *Label Malay equivalent*<br>        Malay equivalent | **Main part of an entry** |
| *French phrase* (such as compound words)<br>    *English phrase equivalent*<br>    *Malay phrase equivalent* | **Illustrative phrases** |
| *Sub-entry (in French)*<br>    (with the same structure as an entry) | **Sub-entries** |
| *Cross-reference marker*<br>    *Cross-reference entry* | **Cross references** |

Fig. 1: logical form of an entry in the FEM dictionary
(the lines in italic are optional in a entry)

### 3. ASCII normalized external form

A label is linked with every type of information of the logical form and is included in the initial ASCII files. Thus, USM have obtained basic entries such as this given in fig. 2 which corresponds to the french entry "*accident*" (the label 'e' corresponds to the French entry, 'pcn' to pronunciation, 'c' to grammatical category, etc.

```
            e,accident
            pcn,/aksidᵃ/
            c,n.m.
            ee,accident
            me,kemalangan
            p,accident de train/d'avion
            epe,train/plane crash
            mpe,kemalangan keretapi
            mpe,nahas kapal terbang
            sbe,accidenté
            pcn,/aksidᵃte/
            c,a.
            ee,damaged (in an accident)
            me,rosak (dalam kemalangan)
            ee,hurt (in an accident)
            me,tercedera (dalam kemalangan)
            g,(terrain)
            ee,uneven
            me,tidak rata (kawasan, daerah)
            ee,hilly
            me,berbukit
```
Fig. 2: a labelled entry

## II. Goals

We pursue four goals in this project. They are listed below in order of importance.

### 1. Paper formatting

Our first aim is to produce from the ASCII normalized form a paper form of the FEM dictionary with a format approaching that of usual dictionaries (fig. 3). This involves the introduction in the format of fonts, styles, etc.

> **accident** /aksidã/ *n.m.* accident : kemalangan, (kejadian) tidak sengaja, (kejadian) secara kebetulan — **accident de train/d'avion** train/plane crash : kemalangan keretapi, nahas kapal terbang — **accidenté** /aksidãte/ *a.* damaged (in an accident) : rosak (dalam kemalangan) hurt (in an accident) : tercedera (dalam kemalangan) *(terrain)* uneven : tidak rata (kawasan, daerah) hilly : berbukit.

Fig. 3 : an entry of the publishable paper FEM dictionary

### 2. Electronic formatting

We also produce an electronic form. This electronic dictionary is supported by of a generic multilingual dictionary tool, ALEX. The problem is to keep as much as possible of the logical form, so as to allow logical access such as searching on multiple keys, sorting, etc.

### 3. Dictionary revision

"Crossing" of the French-English and English-Malay dictionaries has been made manually by people who were not fluent in French. Thus, some errors remain in both the logical structure and in the content. These errors have to be corrected before producing the final paper form of the FEM dictionary.

### 4. Phonetic codes conversion

USM did not use the standard phonetic transcription (international phonetic alphabet - IPA), but a local transcription using certain characters of the Times™ font, which looks like characters of the IPA. These characters have high ASCII code (128 to 256), thus this rendering is different according to the font. To be portable to PC for instance, the files must contain only lower ASCII characters (32 to 128).

## III. Methodology

Our methodology is generic enough to be applied to other projects dealing with the construction of real-size publishable human-oriented dictionaries. The methodology is based on the use of simple but powerful tools.

### 1. Use of an editor for correcting errors

The problem is to find an appropriate software for this work. The first type of software is databases but our experiences with them (we have used dBASE III and 4D) show that lexicographers don't like to work through DataBase Management Systems. They want to use the same word processor to see the texts they want to index and to construct the dictionary.

The most practical tool would be an editor of structured documents like Grif (André, Furuta & Quint 1989, Phan & Boitet 1992) which can manage the logical form of the dictionary. However, such editors are complex to learn and are not yet available on micros as they require large computing ressources. Hence, we use Word, a widely available commercial word processor.

We approach this notion of structured documents by using Word's "styling" facility. A Word style is a group of paragraph and characters format with a name (e.g. the title of this section has the style '*Title1*' which includes the information about the rendering of this title). We associate a particular style to each logical type of information in the dictionary.

> **accident**
>   /aksidã/
>     n.m.
>       accident
>       **kemalangan**
>       **(kejadian) tidak sengaja**
>       **kejadian) secara kebetulan**
> **accident de train/d'avion**
>       train/plane crash
>       **kemalangan keretapi**
>       **nahas kapal terbang**
> **accidenté**
>   /aksidãte/
>     a.
>       damaged (in an accident)
>       **rosak (dalam kemalangan)**
>       hurt (in an accident)
>       **tercedera (dalam kemalangan)**
>       *(terrain)*
>       uneven
>       **tidak rata (kawasan, daerah)**
>       hilly
>       **berbukit**

Fig. 4 : an entry with styles

### 2. Use of an SLLP for converting formats

To convert the initial normalized ASCII external form (fig. 2) in a printable form (fig. 3), we propose some solutions:

- the first solution is to use Word's macro facility. Unfortunately, that facility is only available on the PC version, and we found it very clumsy to constantly exchange large files between the PC and the Macintosh, not speaking of unexpected character transformation in the phonetic font.
- the second solution is to use transducers, but the commercial transcriptors available are only based on direct correspondences. They cannot take into account a forward context and they generally have no variables (or notion of state). Thus, they are not powerful enough for the problems at hand.

We used LT (Language of Transcriptions), a Specialized Language for Linguistic Programming for writting transcriptors.

LT transducers have one input tape with two reading heads (one standard head and one forward head) and one writing head. They can also handle variables and produce side effects. Thus, this kind of transcriptors is not reversible in general.

There have been previous versions of LT (Lepage 1986) The LT used in our work has been implemented on Macintosh with CLOS (Common Lisp Object System) (Lafourcade 1993) The realization is based mainly on Lisp Macroprogramming on the top of an Automaton Manager.

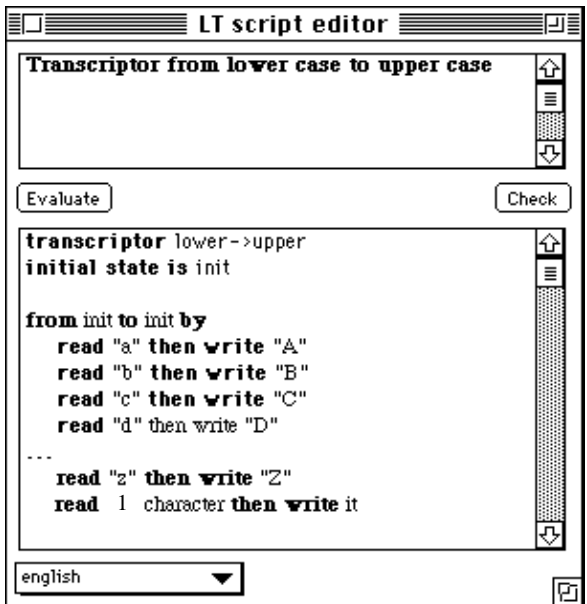

Fig. 5: an LT transcriptor example

With LT, we have easily written all necessary converters.

- Phonetic transcriptions

These conversions first concern the problem of special characters used in some fonts, especially the characters used at USM (standard macintosh fonts, i.e. courier or times) to approximate the international phonetic alphabet (IPA). For example, the ª sign (as in /aksidª/) appears only in a standard macintosh font.

We have thus defined three formats. Ph1 is the initial form of the Word files in a standard macintosh font (files built at USM). Ph2 is the format where special characters are replaced by others which appear in all usual fonts (characters corresponding with the letters, the numbers and with the '+', '-' signs, i.e. 7-bits ASCII). This ASCII coding authorises a safe exchange between Macintoshes and PCs.

    aborigène: /ab>ri3Σn/ → /ABORIJE+N/

To transcript from the Ph1 to the Ph2 formats, we use the LT transcriptor Ph1toPh2. An excerpt is given below.

    **transcriptor** Ph1toPh2
    **initial state is** init

    **from** init **to** init **via**
        read "Σ" **then write** "E+"
        read ">" **then write** "O"
        read "3" **then write** "J"
        read "e" **then write** "E-"
        read "â" **then write** "A-"
        ...
    Fig. 6: excerpt of the LT transcriptor Ph1toPh2

Ph3 is the IPA phonetic format. PalPhon is the font used for this IPA transcription. The problem is to assign this font only to the lines which correspond to phonetic transcription, and hence to determinate the right.

Here, we work on the RTF (Rich Text Format) format, directly produced by Word, which records all the informations describing Word documents (styles, fonts and other informations as italic, bold, etc).

Figure 7 presents the RTF form for the line 'e,accident'. \s0 corresponds to the 'Normal' style of Word and \f20 corresponds to the standard macintosh font.

    \pard\plain   \s0   \f20   e,accident   \par
    Beginning of paragraph        End of paragraph
                Style number
                     Other informations
                        Text of the paragraph
    Fig. 7: description of the RTF

Then, Ph2toPh3 realizes the transition from Ph2 to Ph3. It transforms the RTF form of the lines corresponding to pronunciation (phonetic) by converting the Times font code (\f20) to the PalPhon font code (\f1138) and each character in Ph2 form to the IPA form.

    aborigène: /ABORIJE+N/ —> /abɔriʒɛn/

An excerpt of Ph2toPh3 is given below. The code RTF \'ab corresponds to the character 'ˆ' in the PalPhon font, \'bf to 'ø' , etc.

    **transcriptor** Ph2toPh3
    **initial state is** init

    **from** init **to** init **via**
        read "E+" **then write** "{\f1138 \'ab}"
        read "O" **then write** "{\f1138 \'bf}"
        read "J" **then write** "{\f1138 \'bd}"
        read "E-" **then write** "{\f1138 e}"
        read "A-" **then write** "{\f1138 \'8c}"
        ...
    Fig. 8: excerpt of the LT transcriptor Ph2toPh3

- External format

The first conversion type was about the problem of special characters rendering in the dictionary. The second concerns the external format of the dictionary. We have defined three formats:

- `AN`: the ASCII normalized form which corresponds to the initial files (`Ph1`), these files with phonetic encoding (`Ph2`) and these files in the RTF format (`Ph3`).
- `WT`: the Word transitory form which corresponds to the stylized files with phonetic encoding (`Ph2`) and and these files in the RTF format (`Ph3`) (fig. 4).
- `WP`: the Word printing form (fig. 3) in which we have canceled every informations about styles but we kept the other informations as fonts code and other characters formats (`Ph3`).

The conversions between `AN`, `WT` and `WP` forms are made with LT transcriptors.

### 3. Use of a dictionary tool

Alex is a simple and easy to use generic dictionary tool. Its functionalities are quite classical (inserting and deleting items, sorting, searching). The interesting features are the possibility to index a base on several keys and to search according to these keys or the content of any non-indexed entry (although it is slower).

Entries can be structured objects and searches can be done in function of the values of the features. A same base can handle heterogeneous objects.

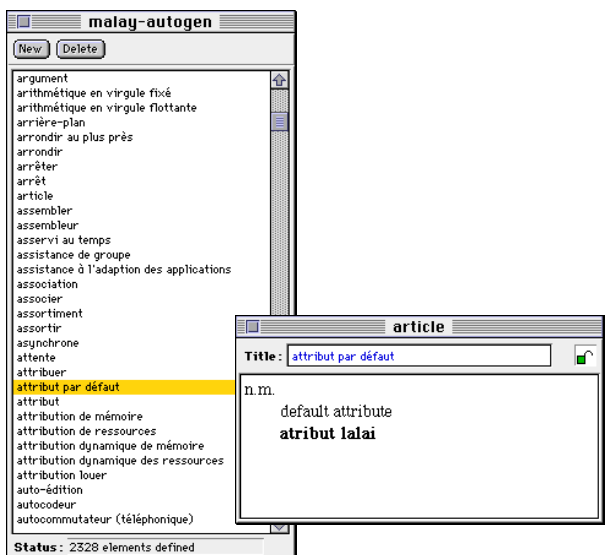

Fig. 9: an example of the electronic form
(terminology entry "attribut par défaut")

It is possible to pilot ALEX remotely (instead of interacting with it via the user-interface) and this method has been used to fill the FEM electronic base.

To do so, we have written an LT transcriptor with strong side effects on ALEX. The goal, here, was not to produce a result in term of a transcripted file, but instead to read a file and produce actions on the ALEX base. As any dialect of LT can mix Lisp commands in their script, it was possible to make these tools cooperate.

## Conclusion

The methodology for manipulating human-oriented dictionaries presented in this paper is based on simple but powerful tools which can be used by lexicographers who don't want to spend much time learning how to use structured documents editors and even less, how to program in DBMS. We use Word, a commercial word processor; LT, a language of transcriptions; ALEX, a dictionary tool. Contrary to our initial fears, these simple tools proved very convenient, and powerful enough for the tasks at hand.

LT and ALEX will soon be available by anonymous ftp at *cambridge.apple.com*.

## Acknowledgements

We are grateful to Chuah C. Kim and Zarin Y. from the University Sains Malaysia for their patience and their corrections of the Malay part of the FEM dictionary. We wish to thank C. Boitet, H. Blanchon, G. Sérasset and other colleagues from GETA, for their support, their help and their remarks. All remaining deficiencies are, of course, ours.

-o-o-o-o-o-o-